\pdfoutput=1

\documentclass[11pt]{article}

\usepackage[review]{acl}

\usepackage{times}
\usepackage{latexsym}

\usepackage[T1]{fontenc}

\usepackage[utf8]{inputenc}

\usepackage{kotex}
\usepackage{multirow}
\usepackage{graphicx}
\usepackage{bm}
\usepackage{amsmath}
\usepackage{amssymb}

\usepackage{microtype}

\title{Paraphrasing via Ranking Many Candidates}

\author{Joosung Lee\\
  Kakao Enterprise Corp., South Korea \\
  \texttt{rung.joo@kakaoenterprise.com} \\}

\begin{document}
\maketitle
\begin{abstract}
We present a simple and effective way to generate a variety of paraphrases and find a good quality paraphrase among them. As in previous studies, it is difficult to ensure that one generation method always generates the best paraphrase in various domains. Therefore, we focus on finding the best candidate from multiple candidates, rather than assuming that there is only one combination of generative models and decoding options. Our approach shows that it is easy to apply in various domains and has sufficiently good performance compared to previous methods. In addition, our approach can be used for data augmentation that extends the downstream corpus, showing that it can help improve performance in English and Korean datasets.
\end{abstract}

\section{Introduction}
Paraphrasing is the task of reconstructing sentences with different words and phrases while maintaining semantic meaning when a source sentence is given. The paraphrase system can be used to add variability to a source sentence and expand it to sentences containing more linguistic information. Paraphrasing has been studied and closely associated with various NLP tasks such as data augmentation, information retrieval, and question answering.

The supervised approach~\cite{patro-etal-2018-learning} to paraphrase is that the model can be trained to generate the paraphrase directly, but requires a parallel dataset. These parallel datasets are expensive to create and difficult to cover various domains. Therefore, in recent years, many studies~\cite{bowman-etal-2016-generating, Miao_Zhou_Mou_Yan_Li_2019, liu-etal-2020-unsupervised} have been conducted on an unsupervised approach to learning paraphrase generation using only the corpus. In addition, there are studies~\cite{mallinson-etal-2017-paraphrasing, thompson-post-2020-paraphrase} that attempt to paraphrase with machine translation learned with a translation corpus (e.g., language pairs shown in WMT~\footnote{http://www.statmt.org/wmt20/}) that has been released widely publicly. Various models have been developed in these methods, but only one model cannot guarantee the best performance for all datasets. Therefore, our goal is not to focus on designing language models or machine translation, but to find best candidates among paraphrases generated by various methods and use them for downstream tasks.

We paraphrase based on a machine translation that can vectorizes sentences with the same meaning in different languages into the same latent representation through an encoder. Our system paraphrases the source sentences with two frameworks and several decoding options and is described in Section~\ref{method_section}. Paraphrase candidates generated in various combinations are ranked according to fluency, diversity, and semantic score. Finally, the system selects a paraphrase that has different words from the source sentence, but is naturally and semantically similar.



The performance and effectiveness of the proposed system are verified in two ways. First, our model is evaluated against a dataset provided with a paraphrase pair. We use QQP (Quora Question Pairs)~\cite{patro-etal-2018-learning} and Medical domain dataset~\cite{medical} and are evaluated by multiple metrics by comparing generated paraphrase and gold reference. The second is to use our system as data augmentation in downstream tasks. We augment financial phrasebank~\cite{Malo2014GoodDO} and hate speech (eng)~\cite{gibert2018hate} in English and hate speech (kor)~\cite{moon-et-al-2020-beep} in Korean to improve the performance of the classification task.

Our system outperforms the previous supervised and unsupervised approaches in terms of the semantic, fluency, and diversity scores shows similar performance to the latest unsupervised approaches. In addition, our system shows performance improvement of downstream tasks, which is a scenario where training data is limited. Finally, our paraphrase has the advantage that it can be applied not only to English but also to various languages.


\section{Methods}
\label{method_section}
\subsection{Pre-trained Model}
We use M2M100~\cite{fan2020beyond} as backbone models so that it can be used not only in English but also in various languages. M2M100 is a multilingual encoder-decoder model that can handle 100 languages, where M2M100-small and M2M100-large two versions are used.

\subsection{Generate Paraphrase Candidates}
We generate paraphrase candidates as follows with two methods according to the combination of encoder and decoder.

\subsubsection{Src-Encoder+Src-Decoder}
\label{src-enc-dec}
The first framework-1 is to use only one language (i.e. source language). Thus, the decoder generates paraphrase candidates directly from the encoded vector of the source sentence. This framework is similar to auto-encoder, but since the paraphrase model is based on a translation system, it has the purpose of generating the same meaning rather than reconstruction.

\subsubsection{Round-trip Translation}
If a candidate sentence is generated with only Section~\ref{src-enc-dec}, the diversity decreases, so the second framework-2 uses two languages to generate more candidates. In other words, we use the round-trip translation mentioned in the \citet{sennrich-etal-2016-improving} to translate the source sentence into the target sentence and translate it back into the source sentence. Because back-translation depends on the performance of the translation system, context information can sometimes be lost, but it can generate various candidates. M2M100 supports 100 languages, but we selected and used English, Korean, French, Japanese, Chinese, German, and Spanish as the language pool.

\subsubsection{Decoder Options}
\label{decoder_options}
When generating paraphrase candidates, we expand the set of candidates by adding various options to the decoder.

In the framework-1, beam search with the beam size of 10 is used and the top-5 candidate sentences are generated. In addition, the following blocking restrictions are additionally applied. (1) Output tokens are restricted so that they do not overlap more than half of the length of the source sentence in succession with the source tokens. (2) It is prevented from generating repetitive 3-grams within the output sentence.

In the framework-2, 3-beam-search is used in both the forward and backward paths, and the top-1 candidate sentence is generated, and the rest are the same as the framework-1.

\subsection{Ranking and Filtering}
\label{ranking_filtering}
We filter through various scores to select the best paraphrase among paraphrase candidates. All ranking and filtering processes measure the score in all lowercase letters to eliminate differences due to uppercase and lowercase letters. The candidates with poor scores in each filtering step are discarded.

\subsubsection{Overlapping}
We remove the overlapping sentences among the candidates that are different from the source sentence. Even in different sentences, candidates that differ only in spaces or by substitution of upper and lower case letters are considered to be the same sentence. The remaining sentences that have been filtered in this section are called $overlap\_cands$.

\subsubsection{Diversity}
\label{diveristy_section}
We measure diversity by comparing $overlap\_cands$ and source sentences. We use word error rate~\cite{MorrisMG04} as diversity metrics, where the higher the score, the higher the diversity. WER (word error rate) refers to the Levenshtein distance between the source sentence and the candidates, and works at the word level instead of the phoneme level. Originally, WER was proposed to measure the performance of an automatic speech recognition system, but we use it to measure the difference between sentences. In this step, only min(5, \#num($overlap\_cands$)/2) sentences with a high diversity score are left, and this is called $diversity\_cands$.

\subsubsection{Fluency}
\label{fluency_section}
To evaluate fluency, we measure PPL (perplexity) using a language model. Fluency indicates the naturalness of the sentence, and the lower the PPL, the better the fluency. We use GPT2-medium~\cite{radford2019language} as the language model and leave only min(3, \#num($diversity\_cands$)/2) sentences with a low PPL, and call this $fluency\_cands$.

\subsubsection{Semantic}
Semantic score measures using a bidirectional pre-trained language model. BERTScore~\cite{Zhang*2020BERTScore} leverages the contextual embeddings and matches words in the candidates and the source sentence by cosine similarity. Higher scores mean semantic similarity, and we use RoBERTa-large~\cite{liu2020roberta} in BERTScore. We measure the semantic score using the source sentence as a reference and $fluency\_cands$ as candidates.


\subsection{Details}
If the source sentence is very short or given a simple structure, in order to obtain more candidates, the decoder options in Section~\ref{decoder_options} are restricted so that the source and output sentences do not overlap more than 2-grams. 

\begin{table}[]
\centering
\resizebox{0.7\columnwidth}{!}{
\begin{tabular}{|c|c|c|c|}
\hline
Dataset            & train & dev & test \\ \hline\hline
Financial Phrasebank         & 1834  & 203 & 227  \\ \hline
Hate   Speech (eng) & 1081  & 220 & 255  \\ \hline
Hate   Speech (kor) & 1421  & 789 & 471  \\ \hline
\end{tabular}
}
\caption{Downstream Datasets}
\label{Tab:donwstream_datasets}
\end{table}

\section{Experiments}
Our training and tests are tested on a single V100 GPU, and the details are described in this Section.

\subsection{Paraphrasing}
\subsubsection{Dataset}
To measure the performance of paraphrase systems, we used Quora Question Pairs (QQP) test data with 30,000 pairs used in ~\citet{patro-etal-2018-learning} and medical domain dataset~\cite{medical}. 

\subsubsection{Evaluation Metrics}
We measure the semantic, diversity, and fluency scores of paraphrases. To set Section~\ref{ranking_filtering} and the evaluation metric differently, diversity uses Isacreblue (inverser-sacrebleu). Isacrebleu is calculated as 100-sacrebleu~\cite{post-2018-call}, and the higher the number of overlapping n-grams between candidates and source sentences, the lower the score.
The semantic score is measured by comparing it with the gold references provided by the dataset and using Bleurt~\cite{sellam-etal-2020-bleurt}. Bleurt is an evaluation metric trained on biased training data so that BERT can model human judgments. We use bleurt-base-128 as the model for Bleurt. When measuring Fluency, GPT2-small is used as a language model.

\subsection{Downstream Task}
To demonstrate the usefulness of our approach, we paraphrase several downstream datasets to experiment with the effects of data augmentation. We test sentence classification in the domains of financial phrasebank~\cite{Malo2014GoodDO} and hate speech~\cite{gibert2018hate} to check usefulness in various domains. It is also paraphrased in hate speech~\cite{moon-et-al-2020-beep} in Korean to check its usefulness not only in English but also in other languages.

\begin{table*}[!t]
\centering
\resizebox{1.45\columnwidth}{!}{
\begin{tabular}{|c|c|c|c|c|c|c|c|}
\hline
\multicolumn{2}{|c|}{\multirow{3}{*}{Methods}} & \multicolumn{3}{c|}{QQP}                           & \multicolumn{3}{c|}{Medical}                        \\ \cline{3-8} 
\multicolumn{2}{|c|}{}                         & Semantic       & Diversity       & Fluency         & Semantic        & Diversity       & Fluency         \\ \cline{3-8} 
\multicolumn{2}{|c|}{}                         & Bleurt         & isacrebleu      & PPL             & Bleurt          & isacrebleu      & PPL             \\ \hline\hline
\multirow{2}{*}{supervised}   & Edlp           & -1.066         & \textbf{86.843} & 585.384         & -               & -               & -               \\ \cline{2-8} 
                              & Edlps          & -0.857         & 83.504          & 597.024         & -               & -               & -               \\ \hline
\multirow{4}{*}{unsuperivsed} & UPSA           & -0.729         & 65.749          & 392.833         & -1.351          & \textbf{89.418} & 476.069         \\ \cline{2-8} 
                              & CGMH(50)       & -0.842         & 65.35           & 556.163         & -1.405          & 88.95           & 818.307         \\ \cline{2-8} 
                              & M2M100         & 0.036          & 43.539          & 346.17          & -0.561          & 35.688          & 296.672         \\ \cline{2-8} 
                              & Ours           & \textbf{0.083} & 69.421          & \textbf{171.61} & \textbf{-0.508} & 68.735          & \textbf{158.76} \\ \hline
\multirow{2}{*}{source}       & input sentence & 0.124          & 0               & 270.781         & -0.523          & 0               & 249.107         \\ \cline{2-8} 
                              & gold reference & 1              & 72.002          & 278.163         & 1               & 88.632          & 171.786         \\ \hline
\end{tabular}
}
\caption{Paraphrasing performance of our approach and previous studies in QQP and Medical. The parentheses of CGMH mean iteration in which the sentence is modified with sample time. Bold text means the best performance.}
\label{Tab:parphrasing_results}
\end{table*}

We download the datasets using huggingface's dataset library~\footnote{https://huggingface.co/datasets/\{financial\_phrasebank, hate\_speech18, kor\_hate\}}. Financial phrasebank and hate speech (eng) are randomly divided into training, validation, and test data because only training data is provided. Hate speech (kor) provides training and test data, so a portion of the training data is used as validation. Since our purpose is to confirm the performance improvement with data augmented by paraphrase in a scenario where there is insufficient data, we preprocess hate speech as follows. (1) In hate speech (eng), the data class is unbalanced, so the data of the class that appears excessively is discarded at random to balance the data. Also, since the amount of existing training data is sufficiently large, in order to limit it to a scenario where data is insufficient, we only use 50\% of the randomly balanced training data. (2) Hate speech (kor) similarly has enough training data, so only 20\% of the training data is randomly used for training. Table~\ref{Tab:donwstream_datasets} shows the statistics of the processed downstream tasks and the performance is measured by accuracy.

\section{Results}
\subsection{Paraphrasing}
\label{paraphrasing_sec}
Table~\ref{Tab:parphrasing_results} shows the performance of paraphrase. \textbf{Edlp} and \textbf{Edlps} are supervised learning models introduced in~\citet{patro-etal-2018-learning}, ED, L, P and S stand for encoder-decoder, cross-entropy, pair-wise discriminator loss, and parameter sharing, respectively. \textbf{CGMH}~\cite{Miao_Zhou_Mou_Yan_Li_2019} uses Metropolis-Hastings sampling in word space to generate constrained sentences. \textbf{UPSA}~\cite{liu-etal-2020-unsupervised} is a method of generating Unsupervised Paraphrase through Simulated Annealing, which searches the sentence space towards this objective by performing a sequence of local edits. \textbf{M2M100} is an M2M-large model that paraphrases source sentences with greedy search (top-1) in framework-1.

Our approach achieves the best performance in terms of semantic and fluency scores than previous studies of supervised and unsupervised methods. The diversity score is not the best performance, but it achieves a score comparable to other models. M2M100, which generates a paraphrase using the same model as ours, achieves the second semantic score, but the diversity is worse than the previous methods. That is, the method of generating simply as a translation model as one option is not perfect, and the rate of generating by copying source sentences from M2M100 in the QQP dataset is 8.41\%.

\subsection{Downstream Task}
Table~\ref{Tab:downstream_results} shows the performance of sentence classification, which are downstream tasks. BERT-base is a bidirectional pre-trained language model. Transformer has the same architecture, but trains from scratch. Both models are trained five times and are the average of the measured performances. We observe that the performance of models is improved when the augmented corpus is used for training.

Because BERT is a pre-trained language model trained from numerous corpuses, it has the ability to extract contextual knowledge. Nevertheless, adding the corpus augmented with paraphrase improves the performance, which shows that it helps training even when fine-tuning the pre-trained language model. Transformers trained from scratch do not have general knowledge of the language, so performance changes through data augmentation are large. Performance is greatly improved in financial and hate speech (eng), but data augmentation in Transformer degrades performance in hate speech (kor). We find that Transformer can learn rich representations through paraphrasing of training data, but performance degradation can occur on fixed test data with a small amount of data. 

Data augmentation through M2M also shows a similar pattern to ours, but the performance improvement is small and the performance degradation is large. We infer that, as shown in Section~\ref{paraphrasing_sec}, the paraphrase performance difference and M2M generate some overlapping sentences. 


\begin{table}[!t]
\centering
\resizebox{1.0\columnwidth}{!}{
\begin{tabular}{|c|c|c|c|c|}
\hline
Methods                      & augmentation & Financial      & \begin{tabular}[c]{@{}c@{}}Hate   Speech\\      (eng)\end{tabular} & \begin{tabular}[c]{@{}c@{}}Hate   Speech\\      (kor)\end{tabular} \\ \hline\hline
\multirow{3}{*}{BERT-base}   & x            & 95.3           & 64.94                                                              & 52.78                                                              \\ \cline{2-5} 
                             & M2M          & 95.15          & 66.2                                                               & 54.52                                                              \\ \cline{2-5} 
                             & Ours         & \textbf{96.33} & \textbf{68.31}                                                     & \textbf{55.03}                                                     \\ \hline
\multirow{3}{*}{Transformer} & x            & 80.47          & 53.24                                                              & \textbf{52.27}                                                     \\ \cline{2-5} 
                             & M2M          & 85.9           & 55.69                                                              & 49.26                                                              \\ \cline{2-5} 
                             & Ours         & \textbf{86.49} & \textbf{63.14}                                                     & 51.04                                                              \\ \hline
\end{tabular}
}
\caption{Accuracy of fine-tuned models in downstream tasks. The performance of each model is the average of the values measured by experimenting five times.}
\label{Tab:downstream_results}
\end{table}

\section{Conclusion}
We propose a system that generates various paraphrase candidates and finds the best candidate through multiple scores, which avoids the risk of relying on one model and one decoding option. Our approach captures semantic information better than the previous supervised and unsupervised methods and generates more natural sentences. The diversity score also achieves similar performance to the state-of-the-art unsupervised method. However, our approach may suffer from speed issues for inferencing heavy models in parallel on one server. For actual paraphrase use, it will be effective to extract candidates along with a simple model such as n-gram.

Our system shows that when data is insufficient in various domains, the classification performance can be improved through data augmentation through our paraphrasing. Our approach is easily extensible across many domains and languages, and we hope to help with a variety of NLP tasks, such as classification tasks with little data.

\bibliography{anthology,custom}

\begin{thebibliography}{18}
\expandafter\ifx\csname natexlab\endcsname\relax\def\natexlab#1{#1}\fi

\bibitem[{Bowman et~al.(2016)Bowman, Vilnis, Vinyals, Dai, Jozefowicz, and
  Bengio}]{bowman-etal-2016-generating}
Samuel~R. Bowman, Luke Vilnis, Oriol Vinyals, Andrew Dai, Rafal Jozefowicz, and
  Samy Bengio. 2016.
\newblock \href {https://doi.org/10.18653/v1/K16-1002} {Generating sentences
  from a continuous space}.
\newblock In \emph{Proceedings of The 20th {SIGNLL} Conference on Computational
  Natural Language Learning}, pages 10--21, Berlin, Germany. Association for
  Computational Linguistics.

\bibitem[{de~Gibert et~al.(2018)de~Gibert, Perez, Garc{\'\i}a-Pablos, and
  Cuadros}]{gibert2018hate}
Ona de~Gibert, Naiara Perez, Aitor Garc{\'\i}a-Pablos, and Montse Cuadros.
  2018.
\newblock \href {https://doi.org/10.18653/v1/W18-5102} {{Hate Speech Dataset
  from a White Supremacy Forum}}.
\newblock In \emph{Proceedings of the 2nd Workshop on Abusive Language Online
  ({ALW}2)}, pages 11--20, Brussels, Belgium. Association for Computational
  Linguistics.

\bibitem[{Fan et~al.(2020)Fan, Bhosale, Schwenk, Ma, El-Kishky, Goyal, Baines,
  Celebi, Wenzek, Chaudhary et~al.}]{fan2020beyond}
Angela Fan, Shruti Bhosale, Holger Schwenk, Zhiyi Ma, Ahmed El-Kishky,
  Siddharth Goyal, Mandeep Baines, Onur Celebi, Guillaume Wenzek, Vishrav
  Chaudhary, et~al. 2020.
\newblock Beyond english-centric multilingual machine translation.
\newblock \emph{arXiv preprint arXiv:2010.11125}.

\bibitem[{Liu et~al.(2020{\natexlab{a}})Liu, Mou, Meng, Zhou, Zhou, and
  Song}]{liu-etal-2020-unsupervised}
Xianggen Liu, Lili Mou, Fandong Meng, Hao Zhou, Jie Zhou, and Sen Song.
  2020{\natexlab{a}}.
\newblock \href {https://doi.org/10.18653/v1/2020.acl-main.28} {Unsupervised
  paraphrasing by simulated annealing}.
\newblock In \emph{Proceedings of the 58th Annual Meeting of the Association
  for Computational Linguistics}, pages 302--312, Online. Association for
  Computational Linguistics.

\bibitem[{Liu et~al.(2020{\natexlab{b}})Liu, Ott, Goyal, Du, Joshi, Chen, Levy,
  Lewis, Zettlemoyer, and Stoyanov}]{liu2020roberta}
Yinhan Liu, Myle Ott, Naman Goyal, Jingfei Du, Mandar Joshi, Danqi Chen, Omer
  Levy, Mike Lewis, Luke Zettlemoyer, and Veselin Stoyanov. 2020{\natexlab{b}}.
\newblock \href {https://openreview.net/forum?id=SyxS0T4tvS} {Ro{\{}bert{\}}a:
  A robustly optimized {\{}bert{\}} pretraining approach}.

\bibitem[{Mallinson et~al.(2017)Mallinson, Sennrich, and
  Lapata}]{mallinson-etal-2017-paraphrasing}
Jonathan Mallinson, Rico Sennrich, and Mirella Lapata. 2017.
\newblock \href {https://www.aclweb.org/anthology/E17-1083} {Paraphrasing
  revisited with neural machine translation}.
\newblock In \emph{Proceedings of the 15th Conference of the {E}uropean Chapter
  of the Association for Computational Linguistics: Volume 1, Long Papers},
  pages 881--893, Valencia, Spain. Association for Computational Linguistics.

\bibitem[{Malo et~al.(2014)Malo, Sinha, Korhonen, Wallenius, and
  Takala}]{Malo2014GoodDO}
P.~Malo, A.~Sinha, P.~Korhonen, J.~Wallenius, and P.~Takala. 2014.
\newblock Good debt or bad debt: Detecting semantic orientations in economic
  texts.
\newblock \emph{Journal of the Association for Information Science and
  Technology}, 65.

\bibitem[{McCreery et~al.(2020)McCreery, Katariya, Kannan, Chablani, and
  Amatriain}]{medical}
Clara~H. McCreery, Namit Katariya, Anitha Kannan, Manish Chablani, and Xavier
  Amatriain. 2020.
\newblock \href {https://doi.org/10.1145/3394486.3412861} {Effective transfer
  learning for identifying similar questions: Matching user questions to
  covid-19 faqs}.
\newblock In \emph{Proceedings of the 26th ACM SIGKDD International Conference
  on Knowledge Discovery \&; Data Mining}, KDD '20, page 3458–3465, New York,
  NY, USA. Association for Computing Machinery.

\bibitem[{Miao et~al.(2019)Miao, Zhou, Mou, Yan, and
  Li}]{Miao_Zhou_Mou_Yan_Li_2019}
Ning Miao, Hao Zhou, Lili Mou, Rui Yan, and Lei Li. 2019.
\newblock \href {https://doi.org/10.1609/aaai.v33i01.33016834} {Cgmh:
  Constrained sentence generation by metropolis-hastings sampling}.
\newblock \emph{Proceedings of the AAAI Conference on Artificial Intelligence},
  33(01):6834--6842.

\bibitem[{Moon et~al.(2020)Moon, Cho, and Lee}]{moon-et-al-2020-beep}
Jihyung Moon, Won~Ik Cho, and Junbum Lee. 2020.
\newblock \href {https://www.aclweb.org/anthology/2020.socialnlp-1.4} {{BEEP}!
  {K}orean corpus of online news comments for toxic speech detection}.
\newblock In \emph{Proceedings of the Eighth International Workshop on Natural
  Language Processing for Social Media}, pages 25--31, Online. Association for
  Computational Linguistics.

\bibitem[{Morris et~al.(2004)Morris, Maier, and Green}]{MorrisMG04}
A.~Morris, V.~Maier, and P.~Green. 2004.
\newblock From wer and ril to mer and wil: improved evaluation measures for
  connected speech recognition.
\newblock In \emph{INTERSPEECH}.

\bibitem[{Patro et~al.(2018)Patro, Kurmi, Kumar, and
  Namboodiri}]{patro-etal-2018-learning}
Badri~Narayana Patro, Vinod~Kumar Kurmi, Sandeep Kumar, and Vinay Namboodiri.
  2018.
\newblock \href {https://www.aclweb.org/anthology/C18-1230} {Learning semantic
  sentence embeddings using sequential pair-wise discriminator}.
\newblock In \emph{Proceedings of the 27th International Conference on
  Computational Linguistics}, pages 2715--2729, Santa Fe, New Mexico, USA.
  Association for Computational Linguistics.

\bibitem[{Post(2018)}]{post-2018-call}
Matt Post. 2018.
\newblock \href {https://www.aclweb.org/anthology/W18-6319} {A call for clarity
  in reporting {BLEU} scores}.
\newblock In \emph{Proceedings of the Third Conference on Machine Translation:
  Research Papers}, pages 186--191, Belgium, Brussels. Association for
  Computational Linguistics.

\bibitem[{Radford et~al.(2019)Radford, Wu, Child, Luan, Amodei, and
  Sutskever}]{radford2019language}
Alec Radford, Jeffrey Wu, Rewon Child, David Luan, Dario Amodei, and Ilya
  Sutskever. 2019.
\newblock Language models are unsupervised multitask learners.
\newblock \emph{OpenAI blog}, 1(8):9.

\bibitem[{Sellam et~al.(2020)Sellam, Das, and Parikh}]{sellam-etal-2020-bleurt}
Thibault Sellam, Dipanjan Das, and Ankur Parikh. 2020.
\newblock \href {https://doi.org/10.18653/v1/2020.acl-main.704} {{BLEURT}:
  Learning robust metrics for text generation}.
\newblock In \emph{Proceedings of the 58th Annual Meeting of the Association
  for Computational Linguistics}, pages 7881--7892, Online. Association for
  Computational Linguistics.

\bibitem[{Sennrich et~al.(2016)Sennrich, Haddow, and
  Birch}]{sennrich-etal-2016-improving}
Rico Sennrich, Barry Haddow, and Alexandra Birch. 2016.
\newblock \href {https://doi.org/10.18653/v1/P16-1009} {Improving neural
  machine translation models with monolingual data}.
\newblock In \emph{Proceedings of the 54th Annual Meeting of the Association
  for Computational Linguistics (Volume 1: Long Papers)}, pages 86--96, Berlin,
  Germany. Association for Computational Linguistics.

\bibitem[{Thompson and Post(2020)}]{thompson-post-2020-paraphrase}
Brian Thompson and Matt Post. 2020.
\newblock \href {https://www.aclweb.org/anthology/2020.wmt-1.67} {Paraphrase
  generation as zero-shot multilingual translation: Disentangling semantic
  similarity from lexical and syntactic diversity}.
\newblock In \emph{Proceedings of the Fifth Conference on Machine Translation},
  pages 561--570, Online. Association for Computational Linguistics.

\bibitem[{Zhang* et~al.(2020)Zhang*, Kishore*, Wu*, Weinberger, and
  Artzi}]{Zhang*2020BERTScore}
Tianyi Zhang*, Varsha Kishore*, Felix Wu*, Kilian~Q. Weinberger, and Yoav
  Artzi. 2020.
\newblock \href {https://openreview.net/forum?id=SkeHuCVFDr} {Bertscore:
  Evaluating text generation with bert}.
\newblock In \emph{International Conference on Learning Representations}.

\end{thebibliography}
\bibliographystyle{acl_natbib}


\end{document}